\documentclass{article}

\usepackage{arXiv}
\usepackage[utf8]{inputenc} % allow utf-8 input
\usepackage[T1]{fontenc}    % use 8-bit T1 fonts
\usepackage{hyperref}       % hyperlinks
\usepackage{url}            % simple URL typesetting
\usepackage{booktabs}       % professional-quality tables
\usepackage{amsfonts}       % blackboard math symbols
\usepackage{nicefrac}       % compact symbols for 1/2, etc.
\usepackage{microtype}      % microtypography
\usepackage{xcolor}         % colors
\usepackage{float}
\usepackage{caption}
\usepackage{graphicx}
\usepackage{multirow}
\usepackage{subcaption}
\usepackage{enumitem, array}
\usepackage{amsmath}
\usepackage{pifont}
\title{Channel-Wise Contrastive Learning for Learning with Noisy Labels}

\author{%
  Hui Kang$^1$ \quad
  Sheng Liu$^2$\thanks{Equal contribution.} \quad
  Huaxi Huang$^3$\footnotemark[1] \quad
  Tongliang Liu$^1$\thanks{Contact person (tongliang.liu@sydney.edu.au).} \\
  $^1$The University of Sydney \quad
  $^2$New York University \quad
  $^3$Data61, CSIRO
}

\begin{document}

\maketitle

\begin{abstract}
In real-world datasets, noisy labels are pervasive. The challenge of learning with noisy labels (LNL) is to train a classifier that discerns the actual classes from given instances. For this, the model must identify features indicative of the authentic labels. While research indicates that genuine label information is embedded in the learned features of even inaccurately labeled data, it's often intertwined with noise, complicating its direct application. Addressing this, we introduce channel-wise contrastive learning (CWCL). This method distinguishes authentic label information from noise by undertaking contrastive learning across diverse channels. Unlike conventional instance-wise contrastive learning (IWCL), CWCL tends to yield more nuanced and resilient features aligned with the authentic labels. Our strategy is twofold: firstly, using CWCL to extract pertinent features to identify cleanly labeled samples, and secondly, progressively fine-tuning using these samples. Evaluations on several benchmark datasets validate our method's superiority over existing approaches.
\end{abstract}

\section{Introduction}
Deep neural networks (DNNs) have been at the forefront of numerous breakthroughs in diverse areas of study, showcasing exemplary performances across a myriad of tasks \cite{he2016deep,girshick2014rich,ronneberger2015u,devlin2018bert}. A significant portion of their success can be attributed to the availability of vast volumes of high-quality annotated data. However, obtaining such expansive and impeccably labeled datasets can often pose significant financial constraints or, in some scenarios, may be entirely unfeasible.

Compounding this challenge, several datasets, as highlighted by sources like \cite{song2019selfie,xiao2015learning}, are amassed through avenues such as search engines or web crawlers. This modus operandi inevitably introduces a plethora of noisy labels. Training DNNs on these compromised datasets can lead to a counterproductive phenomenon: due to the intricate model capacity of DNNs, they tend to adapt excessively to these erroneous labels. This maladaptation subsequently results in compromised model generalization.

The paradigm of learning with noisy labels (LNL) \cite{angluin1988learning} poses intricate challenges. The end goal is to sculpt a classifier that remains resilient against the pitfalls of misleading data emerging from inaccurate labels. Moreover, a successful model would need to deduce the genuine labels predicated upon the features gleaned from the input data. Achieving this necessitates that the derived features are predominantly infused with precise label information. This underscores the critical nature of directing the model to extract and prioritize clean label data in LNL scenarios. The research community has been relentless in this pursuit, exploring an array of methodologies \cite{patrini2017making,zhang2018generalized,han2018co,harutyunyan2020improving,ma2020normalized} to distill pure label information from the mire of noisy training data.

Delving deeper into the anatomy of a DNN, it becomes apparent that the channels of its deep features often resonate with distinct visual patterns \cite{simon2015neural,zhang2016picking,zheng2017learning}. A plethora of studies \cite{chen2017sca,chang2020devil,NEURIPS2019_959ef477,gao2020channel} reinforce the idea that intricate local features — like the unique characteristics of a bird's tail — are sequestered within specific channels. These features play pivotal roles in distinguishing between intricate subjects, such as different avian species. While these channels are treasure troves of clean label information, the full potential of this channel-wise data remains untapped in the context of LNL. The quest to devise strategies that can effectively mine these channels within noisy datasets stands as a tantalizing research challenge.

Recent innovations in the realm of self-supervised learning \cite{chen2020simple,zbontar2021barlow,he2020momentum} have imbued optimism. These methodologies have demonstrated their prowess in amplifying the performance of DNNs, even in the absence of labels. By employing a contrastive loss mechanism on an unlabeled dataset, these self-supervised models have managed to rival, if not surpass, the efficacy of their supervised counterparts that train on labeled data. This observation elucidates the sheer potency of contrastive learning as a tool to unearth the pristine label information embedded within training data.

One underlying culprit behind DNNs' propensity to overfit to label-noise is their reliance on label-dependent supervision losses \cite{arpit2017closer,li2022selective}, such as the cross entropy (CE) loss. This mechanism introduces gradients biased by erroneous labels, marring the optimization process. Recent innovations \cite{yi2022learning,li2022selective} have leveraged the tenets of self-supervised techniques, with contrastive learning being a noteworthy contributor, to counter this overfitting menace. The approach taken by \cite{li2022selective} involves the pre-training of a base model using an unsupervised contrastive loss, followed by the selection of untarnished data pairs for subsequent finetuning. Complementarily, Ref. \cite{yi2022learning} integrated contrastive loss as a complementary regularizer for the CE loss, aiming to achieve robust LNL representation learning. Intriguingly, both these methods \cite{yi2022learning,li2022selective} deploy instance-wise contrastive learning (IWCL) mechanisms to delineate between diverse data instances.

However, while IWCL offers respite from the overfitting challenge by bypassing label dependency, it does not inherently cater to the meticulous extraction of features pivotal to genuine labels. This realization paves the way for the potential inception of an avant-garde self-supervised learning paradigm. It should be meticulously crafted to extensively sift through the data and harvest authentic label information. With this vision, we put forth our innovative channel-wise contrastive learning (CWCL) approach. This technique is tailor-made to hone in on fine-grained and resilient features, setting a new benchmark for LNL scenarios.

\begin{figure}[t]
\begin{center}
\captionsetup[subfigure]{labelformat=empty}
\begin{subfigure}{.118\linewidth}
\includegraphics[width=\linewidth]{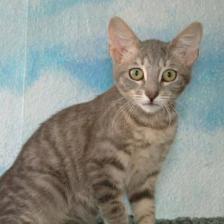}
\end{subfigure}
\begin{subfigure}{.118\linewidth}
\includegraphics[width=\linewidth]{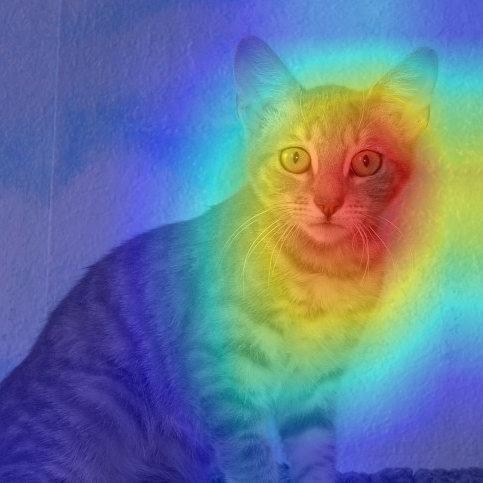}
\end{subfigure}
\begin{subfigure}{.118\linewidth}
\includegraphics[width=\linewidth]{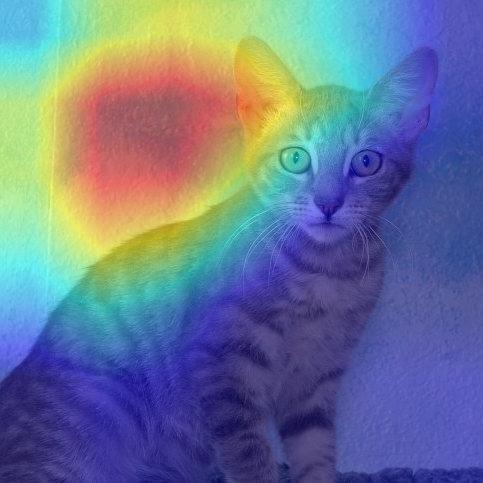}
\end{subfigure}
\begin{subfigure}{.118\linewidth}
\includegraphics[width=\linewidth]{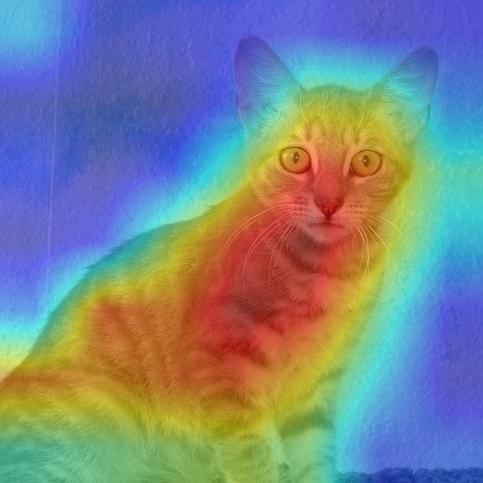}
\end{subfigure}
\hspace{1pt}
\begin{subfigure}{.118\linewidth}
\includegraphics[width=\linewidth]{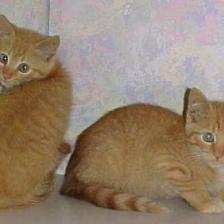}
\end{subfigure}
\begin{subfigure}{.118\linewidth}
\includegraphics[width=\linewidth]{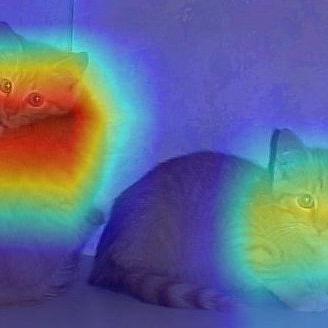}
\end{subfigure}
\begin{subfigure}{.118\linewidth}
\includegraphics[width=\linewidth]{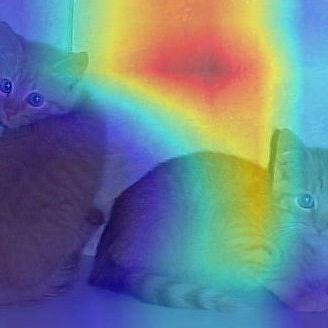}
\end{subfigure}
\begin{subfigure}{.118\linewidth}
\includegraphics[width=\linewidth]{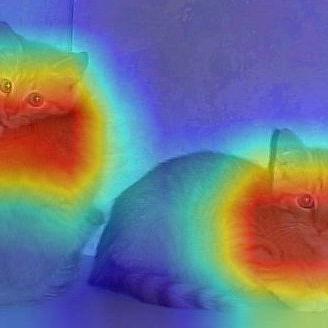}
\end{subfigure}

\vspace{1pt}
\begin{subfigure}{.118\linewidth}
\includegraphics[width=\linewidth]{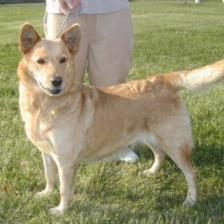}
\caption{Original}
\end{subfigure}
\begin{subfigure}{.118\linewidth}
\includegraphics[width=\linewidth]{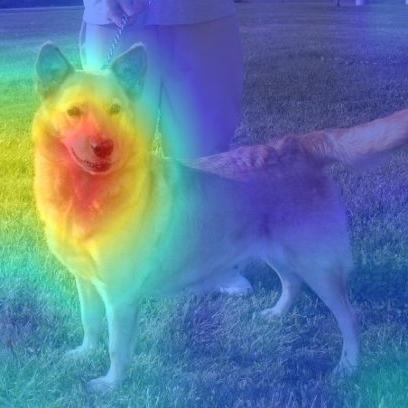}
\caption{Clean}
\end{subfigure}
\begin{subfigure}{.118\linewidth}
\includegraphics[width=\linewidth]{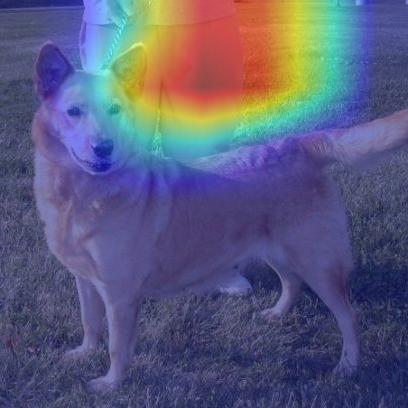}
\caption{Noisy}
\end{subfigure}
\begin{subfigure}{.118\linewidth}
\includegraphics[width=\linewidth]{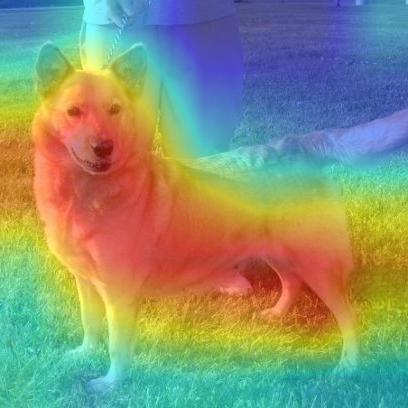}
\caption{Ours}
\end{subfigure}
\hspace{1pt}
\begin{subfigure}{.118\linewidth}
\includegraphics[width=\linewidth]{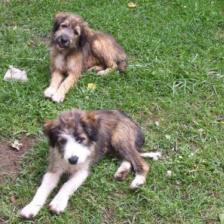}
\caption{Original}
\end{subfigure}
\begin{subfigure}{.118\linewidth}
\includegraphics[width=\linewidth]{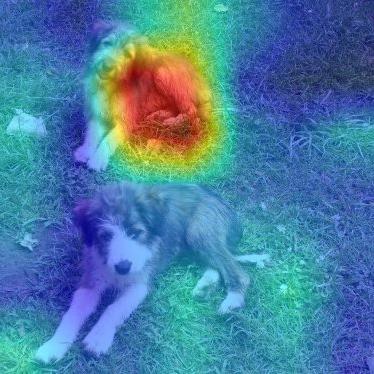}
\caption{Clean}
\end{subfigure}
\begin{subfigure}{.118\linewidth}
\includegraphics[width=\linewidth]{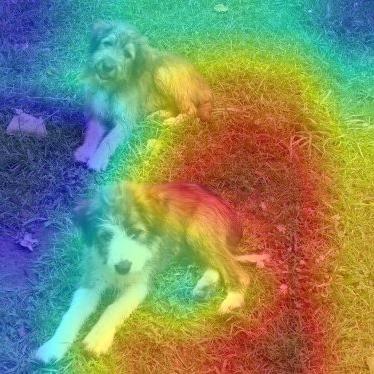}
\caption{Noisy}
\end{subfigure}
\begin{subfigure}{.118\linewidth}
\includegraphics[width=\linewidth]{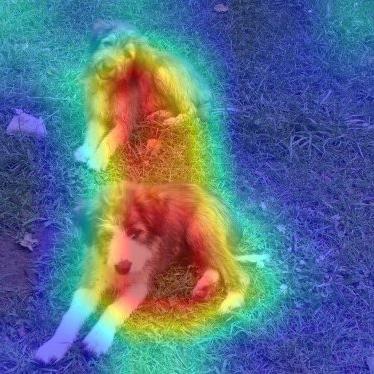}
\caption{Ours}
\end{subfigure}
\end{center}
\caption{The Grad-CAM \cite{selvaraju2017grad} results of ResNet18 models trained on Dogs-vs-Cats \cite{asirra-a-captcha-that-exploits-interest-aligned-manual-image-categorization}. `Clean' represents the model trained with clean labels using CE loss. `Noisy' denotes the model trained with 40\% symmetric label noise using CE loss. `Ours' showcases the model trained with 40\% symmetric label noise using CWCL loss. Under label noise, CWCL can extract similar or even more clean label information than the model trained with clean labels.}
\label{Fig1}
\vspace{-1em}
\end{figure}

Channel-wise contrastive learning is rooted in two fundamental principles: 1) Collaborative training employing CE loss and channel-wise contrastive loss. 2) Progressive confident-sample finetuning. In its operational paradigm, while concurrently trained with CE loss, CWCL delves into individual feature channels, spearheading contrastive learning across every layer or block of the DNN architecture. This unique approach mandates augmented positive channel pairs to gravitate towards each other, whilst ensuring that the negative counterparts maintain discernible distance. A consequential outcome of this mechanism, as illustrated in Figure \ref{Fig1}, is the emergence of a more varied spectrum of feature channels compared to conventional LNL methodologies. 

Enhancing the channel diversity within the corresponding feature maps across layers infuses the input samples or features with profound complexity. This escalating data complexity inversely impacts the model's inherent complexity, ensuring its attenuation. Such a dynamic is pivotal in subduing the model's inclination to overfit when trained on data tainted with noisy labels \footnote{A quintessential learning task demands a harmonious alignment between model and data complexity. Overfitting is inevitable when a highly intricate model grapples with simplistic data, whereas an elementary model runs the risk of underfitting when confronted with convoluted data sets.}. This mechanism acts as a conduit for the LNL model, enabling it to sift and retain pristine label information from a sea of noise. 

Furthermore, drawing from our earlier discussions, the profound depths of feature channels are reservoirs of nuanced and discriminatory data. CWCL capitalizes on this characteristic, meticulously mining these channels to harvest untainted label information. Progressive confident-sample finetuning embarks on a mission to further elevate the prowess of the LNL model. It operates by meticulously curating confident samples from the model's previous training epoch and subsequently leveraging them to nurture the succeeding epoch's model training. This stage is orchestrated through a synergistic training paradigm employing both CE and supervised contrastive loss \cite{khosla2020supervised}. This sequential fine-tuning refines the LNL model incrementally.

To our discernment, CWCL stands as a pioneering methodology that contemplates the challenge of learning features from noisy labels through the lens of a channel-centric perspective. Our pivotal contributions encapsulated within this research paradigm are:
\begin{itemize}[leftmargin=*]
    \item The inception of a channel-wise contrastive loss mechanism, meticulously tailored to extract and amplify clean label information within the realm of the LNL task.
    \item A novel approach towards progressive confident-sample finetuning, designed to incessantly refine the LNL model, thereby enhancing its resilience against label noise.
    \item Our empirical evaluations spanning multiple datasets attest to the formidable efficacy of our proposed strategy. Furthermore, our method consistently eclipses performances set by contemporary state-of-the-art methodologies.
\end{itemize}

\section{Related Work}

\subsection{Learning with Noisy Labels}
The landscape of learning with noisy labels has evolved considerably, and a plethora of methods have emerged. These methodologies can be primarily partitioned into two overarching paradigms: model-based and model-free strategies.

\textbf{Model-based Strategies:} Central to these techniques is the quest to decipher and estimate the intrinsic noise transition probabilities. This is accomplished by characterizing the intricate interplay between the clean and noisy labels \cite{patrini2017making,xia2020robust,xia2020part,yao2020dual,pmlr-v162-liu22w}. Underlying this premise is the assumption that the noisy label manifests as a conditional probability distribution derivative of the pristine labels. Taking illustrative instances, Ref. \cite{goldberger2016training} championed a noise adaptation layer, positioned atop the classification model, dedicated to the learning of transition probabilities. In a nuanced approach, T-revision \cite{xia2019anchor} leveraged fine-tuned slack variables to intuitively gauge the noise transition matrix devoid of anchoring points. Another noteworthy mention is the model proposed by \cite{pmlr-v162-liu22w}, where label noise is encapsulated within a sparse over-parameterized framework.

\textbf{Model-free Strategies:} Distancing themselves from explicit noise modeling, these techniques harness the deep models' inherent memorization capabilities \cite{arpit2017closer} to counteract the deleterious effects of noisy labels \cite{han2018co,li2019dividemix,bai2021understanding,xia2020robust,huang2022paddles}. A case in point is the Co-teaching method \cite{han2018co} where a pair of deep networks embark on a collaborative journey, educating one another through the selective exchange of low-loss instances within mini-batches. Enriching this foundation, DivideMix \cite{li2019dividemix} marries the essence of Co-teaching with the sophistication of two Beta Mixture Models. Further, the PES initiative \cite{bai2021understanding} delves into the intricacies of the progressive early halting of deep architectures, establishing diversified stopping points for distinct network segments. The proposed CWCL method finds its lineage in this model-free territory. It aspires to unearth channel-centric pristine label information via contrastive learning, thereby stifling the negative reverberations of noisy labels during the classification training phase.

\subsection{Feature Channels in DNNs}
A retrospective glance at the annals of DNN research \cite{zeiler2014visualizing,zhou2016learning,selvaraju2017grad} unveils a captivating narrative: network features inherently encode hierarchical information. While the shallower layers resonate with rudimentary semantic nuances like edges and colors, the intermediate strata capture localized or fragmented data. In stark contrast, the pinnacle layers distill high-order semantic facets, painting a comprehensive portrayal of objects. The orchestration of DNNs is such that each layer usually houses multiple filters, each begetting a characteristic feature map. These maps, often envisaged as 2D matrices, amalgamate, forming a 3D tensor when emanating from multiple filters within a singular layer. This assembly culminates in each DNN layer exuding a 3D tensor representation, with individual channels mirroring distinct visual patterns \cite{simon2015neural,zhang2016picking}.

In this realm, several pioneering investigations \cite{chen2017sca,chang2020devil,NEURIPS2019_959ef477,gao2020channel} have spotlighted the profound significance of deep feature channels, particularly in fine-grained image classification endeavors. Their revelations underscore that specific DNN feature channels harbor subtle yet quintessential discriminative part information - a linchpin for intricate image identification. Harnessing such channels promises marked performance leaps in fine-grained image discernment. Drawing inspiration from these insights, we hypothesize that these deep feature channels are reservoirs of untainted label information. Consequently, we champion a novel framework, tailored to extract and leverage these granular discriminative features from channels, catering explicitly to LNL tasks.

\subsection{Contrastive Learning}
Contrastive learning has etched itself as a formidable contender in the representation learning arena \cite{he2020momentum,chen2020simple,caron2020unsupervised,khosla2020supervised}. The core tenet of these methodologies hinges on amplifying the congruence between positive instance pairs while concurrently diluting the affinity between negative counterparts. Incredibly, sans label supervision, contrastive learning has par excellence, rivaling supervised learning across a gamut of tasks.

In the realm of learning with noisy labels, pioneering contributions by \cite{yi2022learning} and \cite{li2022selective} have harnessed the prowess of contrastive learning to counter the overfitting scourge induced by noise-laden labels. To elucidate, Ref. \cite{li2022selective} employed contrastive learning as a precursor to base model pre-training, subsequently cherry-picking pristine pairs for the fine-tuning phase. Contrastingly, Ref. \cite{yi2022learning} wove the contrastive loss into the cross-entropy loss fabric, culminating in robust LNL representation learning. These ventures predominantly orbit around instance-wise contrastive learning, emphasizing differentiating disparate instances. Breaking from this mold, our proposition stands unparalleled as the maiden foray into explicit channel-centric feature learning for LNL. Given that specific feature channels encapsulate fine-grained discriminative nuances, our methodology promises a richer label information extraction than its instance-wise contrastive learning counterparts.

\section{Methodology}
In this section, we first provide the problem definition of a LNL task. Next, we present our proposed channel-wise contrastive learning (CWCL) approach. Finally, we introduce a progressive confident sample fine-tuning technique to further enhance the performance of the LNL classifier.

\subsection{Problem Definition}
In the context of learning with noisy labels, the true distribution of training data is typically represented by $\mathcal{D}= \{ \left({x},y \right)| x \in \mathcal{X}, y \in {1, \dots,K} \}$. Here, $\mathcal{X}$ denotes the sample space, and ${1,\dots,K}$ represents the label space consisting of $K$ classes. However, due to label errors during data collection and dataset construction, the actual distribution of the label space is often unknown. Therefore, we have to rely on a noisy dataset $\mathcal{\tilde{D}}= \{ \left({x},\tilde{y} \right)| x \in \mathcal{X}, \tilde{y} \in {1, \dots,K} \}$ with corrupted labels $\tilde{y}$ to train the model. Our goal is to develop an algorithm that can learn a robust deep classifier from these noisy data to accurately classify query samples. 

\subsection{Channel-Wise Contrastive Loss}
\begin{figure}[ht]
\begin{center}
\captionsetup[subfigure]{labelformat=empty}
\begin{subfigure}{.4\linewidth}
\includegraphics[width=\linewidth]{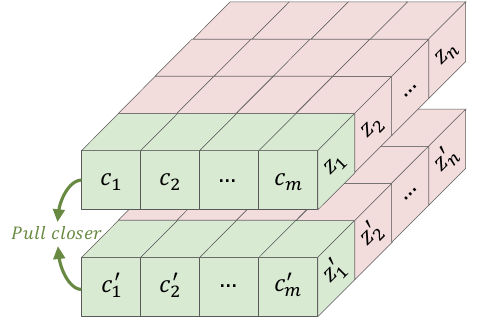}
\caption{Instance-Wise Contrastive}
\end{subfigure}
\begin{subfigure}{.4\linewidth}
\includegraphics[width=\linewidth]{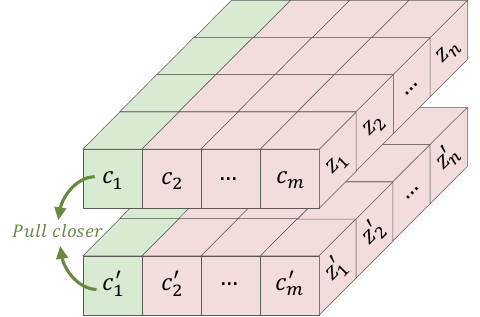}
\caption{Channel-Wise Contrastive}
\end{subfigure}
\end{center}
\caption{Comparison between instance-wise and channel-wise contrastive loss. $z$ is the feature representation for different instances, and $c$ refers to different feature channels. For instance-wise contrastive loss, two views of the same instance, $z_i$ and $z_i^{\prime}$, constitute the positive pair. While for channel-wise contrastive loss, two views of the same channel, $c_i$ and $c_i^{\prime}$, form the positive pair.}
\label{Fig2}
\end{figure}

Following previous traditional instance-wise contrastive learning (IWCL) methods~\cite{chen2020simple,he2020momentum,chen2020improved,chen2020big}, we often perform random data augmentation twice on each image in a mini-batch of $N$ images $\{x_1,x_2,\dots,x_N\}$, resulting in a larger batch of $2N$ images. For convenience, we consider the images $x_i$ and $x_{i+N}$ as two different augmented versions (views) of the same original image, which together form a positive pair. The representation vector from projection head is denoted as $z = Proj(Enc(x))$. The length of $z$ is $M$ (i.e. number of channels). The instance-wise contrastive loss, also known as InfoNCE loss \cite{oord2018representation} is expressed as
\begin{equation}
    \mathcal{L}_{IWCL} = -\sum_{i=1}^{N} \log \frac{\exp\left(sim\left(z_i, z_{i+N}\right)/\tau\right)}{\sum_{k=1, k \neq i}^{2N} \exp\left(sim\left(z_i, z_k\right)/\tau\right)}
    \label{eq1}
\end{equation}
where $\tau$ is a temperature hyper-parameter, and $sim$ represents the cosine similarity. Intuitively, $\mathcal{L}_{IWCL}$ encourages the encoder network to learn similar representation for different augmentations from the same image while increasing the difference between representations of the augmentations from different images.

As for proposed channel-wise contrastive loss, referring to Figure \ref{Fig2} and Equation \ref{eq1}, it can be summarized as
\begin{equation}
    \mathcal{L}_{CWCL} = -\sum_{i=1}^{M} \log \frac{\exp\left(sim\left(c_i, c_{i+M}\right)/\tau\right)}{\sum_{k=1, k \neq i}^{2M} \exp\left(sim\left(c_i, c_k\right)/\tau\right)}
    \label{eq2}
\end{equation}
From Equation \ref{eq2}, we can know that, $\mathcal{L}_{CWCL}$ encourages the encoder network to learn similar representation for different augmentations from the same channel while increasing the difference between representations of the augmentations from different channels.

We extract representations from $L$ intermediate layers of the network (e.g. $layer\{1,\dots,4\}$ of ResNet18) and apply each to $\mathcal{L}_{CWCL}$, and use CE loss ($\mathcal{L}_{CE}$) at the fully connected layer to do classification. So the combined total loss of this stage ($\mathcal{L}_{Stage1}$) is
\begin{equation}\label{stage1}
    \mathcal{L}_{Stage1} = \left(1 - \lambda\right)\mathcal{L}_{CE} + \frac{\lambda}{L}\sum_{l=1}^{L}\mathcal{L}_{CWCL}
\end{equation}
where $\lambda$ is a loss balance hyper-parameter. Training the LNL model using Equation \ref{stage1}, we can obtain a classifier that learns to mine the clean label information from the noisy training data under the supervision of supervised CE loss and unsupervised CWCL loss.

\subsection{Progressive Confident Sample Finetuning}
Confident samples are characterized by high prediction probabilities concerning their associated labels. In a bid to augment robustness, we employ two distinct data augmentations for every input sample. The subsequent predicted label is ascertained by averaging predictions across these augmentations. This strategy not only stabilizes predictions but also elevates performance, as corroborated by empirical studies. On gathering these high-confidence samples, the classifier is then trained by viewing them as pristine, noise-free data.

A noteworthy challenge that surfaces is the potential variance in the number of confident samples across different classes. Training the model directly on this set might propagate severe class imbalance issues. To navigate this predicament, we incorporate a class balance sampler in the dataloader. This ensures an equitable representation from all classes during the model's training phase, thus bolstering its learning efficacy.

Since we consider confident samples as clean data, we use supervised contrastive loss ($\mathcal{L}_{SupCon}$) \cite{khosla2020supervised} instead of $\mathcal{L}_{CWCL}$. Same as $\mathcal{L}_{Stage1}$, we also extract representations from $L$ intermediate layers of the network and apply each to $\mathcal{L}_{SupCon}$, and use CE loss ($\mathcal{L}_{CE}$) at the fully connected layer to do classification. So the combined total loss of this stage ($\mathcal{L}_{Stage2}$) is
\begin{equation} \label{stage2}
    \mathcal{L}_{Stage2} = \left(1 - \lambda\right)\mathcal{L}_{CE} + \frac{\lambda}{L}\sum_{l=1}^{L}\mathcal{L}_{SupCon}
\end{equation}
Based on the confident sample selector trained in stage one using Equation \ref{stage1}, we continuously train the model using Equation \ref{stage2} each time and select the confident sample from the currently trained model as next-time input. Such a progressive confident-sample finetuning strategy can gradually improve the quality of the confident sample and further boost the performance of a LNL model.

\section{Experiment}

\begin{table}[t]
\caption{Comparison of test accuracy using different methods on CIFAR-10 dataset with varying noise types and levels. The baseline results are taken from \cite{yi2022learning}. We use ResNet18 as the architecture, whereas all other methods in the comparison use PreAct ResNet18. The mean and standard deviation over 3 runs are reported. The best results are highlighted in bold.}
\label{Tab1}
\centering
\resizebox{\linewidth}{!}{
\begin{tabular}{c|cccccc}
\hline
\multirow{3}{*}{Method} & \multicolumn{6}{c}{CIFAR-10} \\ \cline{2-7} 
                        & \multicolumn{5}{c|}{Symmetric}                                                                                                                             & Asymmetric                    \\ \cline{2-7} 
                        & 0\%                     & 20\%                    & 40\%                    & 60\%                    & \multicolumn{1}{c|}{80\%}                    & 40\%                    \\ \hline
CE                      & 93.97$\pm$0.22          & $88.51_{\pm0.17}$          & 82.73$\pm$0.16          & 76.26$\pm$0.29          & \multicolumn{1}{c|}{59.25$\pm$1.01}          & 83.23$\pm$0.59          \\
Forward                 & 93.47$\pm$0.19          & 88.87$\pm$0.21          & 83.28$\pm$0.37          & 75.15$\pm$0.73          & \multicolumn{1}{c|}{58.58$\pm$1.05}          & 82.93$\pm$0.74          \\
GCE                     & 92.38$\pm$0.32          & 91.22$\pm$0.25          & 89.26$\pm$0.34          & 85.76$\pm$0.58          & \multicolumn{1}{c|}{70.57$\pm$0.83}          & 82.23$\pm$0.61          \\
Co-teaching             & 93.37$\pm$0.12          & 92.05$\pm$0.15          & 87.73$\pm$0.17          & 85.10$\pm$0.49          & \multicolumn{1}{c|}{44.16$\pm$0.71}          & 77.78$\pm$0.59          \\
LIMIT                   & 93.47$\pm$0.56          & 89.63$\pm$0.42          & 85.39$\pm$0.63          & 78.05$\pm$0.85          & \multicolumn{1}{c|}{58.71$\pm$0.83}          & 83.56$\pm$0.70          \\
SLN                     & 93.21$\pm$0.21          & 88.77$\pm$0.23          & 87.03$\pm$0.70          & 80.57$\pm$0.50          & \multicolumn{1}{c|}{63.99$\pm$0.79}          & 81.02$\pm$0.25          \\
SL                      & 94.21$\pm$0.13          & 92.45$\pm$0.08          & 89.22$\pm$0.08          & 84.63$\pm$0.21          & \multicolumn{1}{c|}{72.59$\pm$0.23}          & 83.58$\pm$0.60          \\
APL                     & 93.97$\pm$0.25          & 92.51$\pm$0.39          & 89.34$\pm$0.33          & 85.01$\pm$0.17          & \multicolumn{1}{c|}{70.52$\pm$2.36}          & 84.06$\pm$0.20          \\
CTRR                    & 94.29$\pm$0.21          & 93.05$\pm$0.32          & 92.16$\pm$0.31          & 87.34$\pm$0.84          & \multicolumn{1}{c|}{83.66$\pm$0.52}          & 89.00$\pm$0.56          \\ \hline
CWCL                    & \textbf{96.71$\pm$0.02} & \textbf{94.04$\pm$0.16} & \textbf{93.46$\pm$0.08} & \textbf{91.87$\pm$0.04} & \multicolumn{1}{c|}{\textbf{86.31$\pm$0.70}} & \textbf{92.71$\pm$0.27} \\ \hline
\end{tabular}}
\end{table}

\begin{table}[t]
\caption{Comparison of test accuracy using different methods on CIFAR-100 dataset with varying noise types and levels. The baseline results are taken from \cite{yi2022learning}. We use ResNet18 as the architecture, whereas all other methods in the comparison use PreAct ResNet18. The mean and standard deviation over 3 runs are reported. The best results are highlighted in bold.}
\label{Tab2}
\centering
\resizebox{\linewidth}{!}{
\begin{tabular}{c|cccccc}
\hline
\multirow{3}{*}{Method} & \multicolumn{6}{c}{CIFAR-100}                                                                                                                                                  \\ \cline{2-7} 
                        & \multicolumn{5}{c|}{Symmetric}                                                                                                                             & Asymmetric                    \\ \cline{2-7} 
                        & 0\%                     & 20\%                    & 40\%                    & 60\%                    & \multicolumn{1}{c|}{80\%}                    & 40\%                    \\ \hline
CE                      & 73.21$\pm$0.14          & 60.57$\pm$0.53          & 52.48$\pm$0.34          & 43.20$\pm$0.21          & \multicolumn{1}{c|}{22.96$\pm$0.84}          & 44.45$\pm$0.37          \\
Forward                 & 73.01$\pm$0.33          & 58.72$\pm$0.54          & 50.10$\pm$0.84          & 39.35$\pm$0.82          & \multicolumn{1}{c|}{17.15$\pm$1.81}          & -                       \\
GCE                     & 72.27$\pm$0.27          & 68.31$\pm$0.34          & 62.25$\pm$0.48          & 53.86$\pm$0.95          & \multicolumn{1}{c|}{19.31$\pm$1.14}          & 46.50$\pm$0.71          \\
Co-teaching             & 73.39$\pm$0.27          & 65.71$\pm$0.20          & 57.64$\pm$0.71          & 31.59$\pm$0.88          & \multicolumn{1}{c|}{15.28$\pm$1.94}          & -                       \\
LIMIT                   & 65.53$\pm$0.91          & 58.02$\pm$1.93          & 49.71$\pm$1.81          & 37.05$\pm$1.39          & \multicolumn{1}{c|}{20.01$\pm$0.11}          & -                       \\
SLN                     & 63.13$\pm$0.21          & 55.35$\pm$1.26          & 51.39$\pm$0.48          & 35.53$\pm$0.58          & \multicolumn{1}{c|}{11.96$\pm$2.03}          & -                       \\
SL                      & 72.44$\pm$0.44          & 66.46$\pm$0.26          & 61.44$\pm$0.23          & 54.17$\pm$1.32          & \multicolumn{1}{c|}{34.22$\pm$1.06}          & 46.12$\pm$0.47          \\
APL                     & 73.88$\pm$0.99          & 68.09$\pm$0.15          & 63.46$\pm$0.17          & 53.63$\pm$0.45          & \multicolumn{1}{c|}{20.00$\pm$2.02}          & 52.80$\pm$0.52          \\
CTRR                    & 74.36$\pm$0.41          & 70.09$\pm$0.45          & 65.32$\pm$0.20          & 54.20$\pm$0.34          & \multicolumn{1}{c|}{43.69$\pm$0.28}          & 54.47$\pm$0.37          \\ \hline
CWCL                    & \textbf{81.04$\pm$0.10} & \textbf{75.12$\pm$0.12} & \textbf{73.90$\pm$0.27} & \textbf{70.83$\pm$0.23} & \multicolumn{1}{c|}{\textbf{60.49$\pm$0.34}} & \textbf{73.97$\pm$0.04} \\ \hline
\end{tabular}}
\end{table}

\subsection{Datasets and Implementation Details}
\textbf{Datasets:}
We evaluate our method on two synthetic datasets with different noise types and levels, CIFAR-10 and CIFAR-100 \cite{glorot2010understanding}, as well as two real-world datasets, Animal-10N \cite{song2019selfie} and Clothing-1M \cite{xiao2015learning}. CIFAR-10 and CIFAR100 both contain 50k training images and 10k testing images, each with a size of 32$\times$32 pixels. CIFAR-10 has 10 classes, while CIFAR-100 contains 100 classes. The original labels of these two datasets are clean. Animal-10N has 10 animal classes with 50k training images and 5k test images, each with a size of 64$\times$64 pixels. Its estimated noise rate is around 8\%. Clothing-1M has 1 million training images and 10k test images with 14 classes crawled from online shopping web sites. Its estimated noise level is around 40\%.

\textbf{Synthetic Noise:}
Following previous works \cite{han2018co,liu2020early,xia2021robust,patrini2017making}, we explore two different types of synthetic noise with different noise levels for both CIFAR-10 and CIFAR-100 datasets. For symmetric label noise in both datasets, each label has the same probability of being flipped to any class, and we randomly select a certain percentage of training data to have their labels flipped, with the range being \{20\%, 40\%, 50\%, 60\%, 80\%\}. For asymmetric label noise in CIFAR-10, we follow the labeling rule proposed in \cite{patrini2017making}, where we flip labels between TRUCK $\rightarrow$ AUTOMOBILE, BIRD $\rightarrow$ AIRPLANE, DEER $\rightarrow$ HORSE, and CAT $\leftrightarrow$ DOG. We randomly choose 40\% of the training data and flip their labels according to the asymmetric labeling rule. For asymmetric label noise in CIFAR-100, we also randomly select 40\% of the training data and flip their labels to the next class in the label space.

\textbf{Baseline Methods:}
To evaluate the effectiveness of our proposed method, we compare it against several baseline methods that address label noise. These methods include:
1) CE loss, which is the standard loss function used in supervised learning.
2) Forward correction \cite{patrini2017making}, which corrects loss values based on an estimated noise transition matrix.
3) GCE \cite{zhang2018generalized}, which combines the Mean Absolute Error (MAE) loss and CE loss to create a robust loss function.
4) Co-teaching \cite{han2018co}, which trains two networks and selects small-loss examples to update.
5) LIMIT \cite{harutyunyan2020improving}, which introduces noise to gradients to avoid memorization.
6) SLN \cite{chen2021noise}, which adds Gaussian noise to labels to combat label noise.
7) SL \cite{wang2019symmetric}, which uses CE loss and a Reverse Cross Entropy (RCE) loss as a robust loss function.
8) APL \cite{ma2020normalized}, which combines two mutually boosted robust loss functions NCE and RCE for training.
9) CTRR \cite{yi2022learning}, which proposes a novel contrastive regularization function to address the memorization issue of LNL, and achieved state-of-the-art performance.
The results of the baseline methods are taken from \cite{yi2022learning}.

\textbf{Implementation Details:}
In our study, to guarantee unbiased evaluations across different experiments, we consistently utilized the ResNet18 architecture \cite{he2016deep} for all datasets. Throughout the training process, we adhered to a batch size of 128 and set the loss balance factor $\lambda$ to 0.6. Optimization was achieved using the SGD optimizer with a momentum of 0.9, weight decay of 5e-4, and an initial learning rate of 0.1. Our training regimen spanned a total of 300 epochs: initially, 100 epochs leveraging the channel-wise contrastive loss were used to form a preliminary model adept at generating high-quality confident samples. These samples subsequently steered the next 200 epochs of training, employing the contrastive deep supervision method as proposed in \cite{zhang2022contrastive}. To fortify the stability of our training dynamics, we adopted the exponential moving average (EMA) strategy \cite{izmailov2018averaging}. Ensuring transparency, all our findings are presented as the mean and standard deviation derived from three independent runs.

\begin{table}[t]
\caption{Comparison of test accuracy using different methods on real-world datasets Animal-10N and Clothing-1M. The baseline results are taken from \cite{yi2022learning}. All methods use ResNet18 as the base model. The mean and standard deviation over 3 runs are reported. The best results are highlighted in bold.}
\label{Tab3}
\centering
\resizebox{0.5\linewidth}{!}{
\begin{tabular}{c|c|c}
\hline
Method      & Animal-10N              & Clothing-1M             \\ \hline
CE          & 83.18$\pm$0.15          & 70.88$\pm$0.45          \\
Forward     & 83.67$\pm$0.31          & 71.23$\pm$0.39          \\
GCE         & 84.42$\pm$0.39          & 71.34$\pm$0.12          \\
Co-teaching & 85.73$\pm$0.27          & 71.68$\pm$0.21          \\
SLN         & 83.17$\pm$0.08          & 71.17$\pm$0.12          \\
SL          & 83.92$\pm$0.28          & 72.03$\pm$0.13          \\
APL         & 84.25$\pm$0.11          & 72.18$\pm$0.21          \\
CTRR        & 86.71$\pm$0.15          & 72.71$\pm$0.19          \\ \hline
CWCL        & \textbf{88.95$\pm$0.23} & \textbf{73.87$\pm$0.16} \\ \hline
\end{tabular}}
\end{table}

\subsection{Classification Performance Analysis}

\textbf{Results on Synthetic Datasets:}
Tables \ref{Tab1} and \ref{Tab2} elucidate the comparative performance of different methods on CIFAR-10 and CIFAR-100 across a diverse spectrum of label noise settings. A discernible trend from the results is the consistent superior performance of CWCL, underscoring its robustness against other contemporary methodologies. While CWCL's dominance is evident across all noise levels, its prowess is particularly pronounced in the CIFAR-100 dataset with symmetric 80\% and asymmetric 40\% noise settings. Here, CWCL surpasses the runner-up method by an impressive margin, nearing 20 percentage points in terms of accuracy. Such outstanding results underscore CWCL's adeptness in managing label noise, thereby ensuring enhanced classification outcomes even under intricate noise-dominant scenarios.

\textbf{Results on Real-world Datasets:}
Table \ref{Tab3} shines a light on the comparative analysis for real-world datasets, namely Animal-10N and Clothing-1M. To ensure an equitable comparison with previously established benchmarks, specific architectures were chosen: a randomly initialized ResNet18 for Animal-10N and an ImageNet pre-trained ResNet18 for Clothing-1M. Even under these settings, CWCL emerges as a top contender, surpassing other methods on both the aforementioned datasets. Its efficacy is especially noteworthy on the Clothing-1M dataset, an expansive collection of 1 million images. On this dataset, CWCL not only establishes its dominance but does so with an appreciable margin, achieving an accuracy that's over a percentage point higher than its closest competitor. This superior performance on a real-world, large-scale dataset further testifies to CWCL's robustness and adaptability.

\subsection{Ablation Studies}

\begin{table}[h]
\caption{Ablation studies of the proposed CWCL under the supervised setting, experiments on CIFAR-100 are based on a ResNet18 backbone.
CWCL($Y$/$N$) indicates the progressive confident-sample finetuning stage is enabled/disabled.}
\label{Tab4}
\centering
\resizebox{0.8\linewidth}{!}{
\begin{tabular}{c|cccc}
\hline
\multirow{3}{*}{Method} & \multicolumn{4}{c}{CIFAR-100}                                                          \\ \cline{2-5} 
                        & \multicolumn{3}{c|}{Symmetric}                                        & Asymmetric     \\ \cline{2-5} 
                        & 40\%           & 60\%           & \multicolumn{1}{c|}{80\%}           & 40\%           \\ \hline
CE                      & 52.48$\pm$0.34 & 43.20$\pm$0.21 & \multicolumn{1}{c|}{22.96$\pm$0.84} & 44.45$\pm$0.37 \\
CTRR                    & 65.32$\pm$0.20 & 54.20$\pm$0.34 & \multicolumn{1}{c|}{43.69$\pm$0.28} & 54.47$\pm$0.37 \\
CWCL$\left(N\right)$    & 70.70$\pm$0.07 & 63.81$\pm$0.26 & \multicolumn{1}{c|}{49.48$\pm$0.40} & 67.91$\pm$0.48 \\
CWCL$\left(Y\right)$    & 73.90$\pm$0.27 & 70.83$\pm$0.23 & \multicolumn{1}{c|}{60.49$\pm$0.34} & 73.97$\pm$0.04 \\ \hline
\end{tabular}}
\end{table}

We present an ablation study to delve deeper into the inner workings and contributions of various components of our proposed CWCL method. The results of this analysis are illustrated in Table \ref{Tab4}.

\textbf{Effectiveness of Channel-Wise Contrastive Learning:} 
By examining the results of CWCL(N), it becomes evident that even without the integration of progressive confident sample finetuning, CWCL outperforms both CE and CTRR. This significant lead highlights the inherent strength and efficacy of the channel-wise contrastive learning approach. By leveraging this scheme, our model efficiently harnesses more discriminative label information, even in the presence of noisy training data, further attesting to its resilience and robustness.

\textbf{Contribution of Progressive Confident Sample Finetuning:}
The results demonstrate that incorporating the progressive confident sample finetuning stage further refines the performance, as seen with CWCL(Y). This observation aligns with our expectations. After the initial training phase via channel-wise contrastive learning, the model possesses the capability to produce high-quality confident samples. These samples then serve as an invaluable resource during subsequent training stages, ensuring a more refined learning process.

\textbf{Performance under High Noise Levels:}
An intriguing observation from our analysis is the enhanced performance of our model, especially when subjected to high noise levels. The progressive confident-sample finetuning technique appears to be particularly effective in these scenarios. This suggests that our method is not only robust but is also adaptive, enhancing its capabilities further when faced with challenging noisy conditions.

In conclusion, our ablation study underscores the synergistic effect of combining channel-wise contrastive learning with progressive confident sample finetuning. Each component plays a pivotal role, contributing to the method's overall effectiveness and resilience against noise.

\section{Conclusion}

In this research, we have introduced the Channel-Wise Contrastive Learning loss (CWCL) - a novel contrastive loss that offers a more refined approach to harnessing true label information, even amid a noisy environment. Unlike traditional methods that might struggle in the presence of noise, our empirical investigations have spotlighted the ability of CWCL to generate noise-resilient, as well as intricately detailed features. These features not only serve as robust indicators of true label information but also pave the way for the selection of pristine, high-caliber samples.
The supremacy of CWCL over traditional instance-wise contrastive learning loss is not merely theoretical. Our empirical results are a testament to its prowess. A salient takeaway from our findings is how CWCL fosters feature diversity, a crucial attribute that bolsters the model's resistance against overfitting by augmenting feature complexity.
However, our exploration did not stop at theoretical datasets. Recognizing the multifaceted nature of real-world data, we subjected CWCL to a series of rigorous tests, spanning a myriad of datasets varying in noise levels, noise types, and even those that mirror the unpredictable nature of real-world label noises. Our method showcased remarkable consistency and effectiveness across all evaluations, further solidifying its practical utility.
Closing this chapter of research, we posit that the scope of CWCL is not limited to operating in isolation. Its adaptable nature makes it a prime candidate for integration with other existing methodologies. In scenarios plagued by label noise, which is a prevalent challenge in many machine learning endeavors, CWCL promises to be a potent tool, either standalone or in conjunction with other techniques, providing an edge in crafting more resilient and accurate models. We are optimistic about its potential and foresee its adoption in numerous forthcoming applications to tackle label noise challenges.

\newpage

{
\small
\bibliography{egbib}
\bibliographystyle{plain}
}

\end{document}